\documentclass[twoside]{article}

\usepackage[utf8]{inputenc}
\usepackage[english]{babel}
\usepackage{pdfpages}
\usepackage{xcolor}
\usepackage[font=it, labelfont =normalfont]{caption}
\usepackage{amsmath}
\usepackage{amsthm}
\usepackage{amsfonts}
\usepackage{amssymb}
\usepackage{mathrsfs}
\usepackage{mathtools}
\usepackage[titletoc]{appendix} 
\usepackage{float}
\usepackage[separate-uncertainty = true]{siunitx}
\usepackage{array}
\usepackage[version=3]{mhchem} 
\usepackage{comment}
\usepackage{enumitem}
\usepackage{booktabs, multirow} % for borders and merged ranges
\newtheorem{theorem}{Theorem}
\usepackage{pdflscape}
\usepackage{afterpage}
\usepackage{tabularx}
\usepackage{longtable}
 \usepackage{url}
 \usepackage{multicol}
\usepackage{hyperref}
\usepackage{xcolor}
\usepackage{microtype}
\usepackage{graphicx}
\usepackage{subfigure}
\usepackage{booktabs} % for professional tables
\usepackage{hyperref}
\usepackage{natbib}
\setcitestyle{authoryear, semicolon, open={(},close={)}}

\usepackage[accepted]{aistats_style}
\usepackage{subfiles}
 
\global\long\def\b#1{\boldsymbol{#1}}

\global\long\def\br#1{\left( #1 \right) }
\global\long\def\sqbr#1{\left[ #1 \right] }

\begin{document}

\twocolumn[

\aistatstitle{Contextual Bayesian optimization with binary outputs}
\aistatsauthor{ Tristan Fauvel \And Matthew Chalk}

\aistatsaddress{Sorbonne Université, INSERM, CNRS, Institut de la Vision \\
    F-75012 Paris, France} ]

\begin{abstract}

Bayesian optimization (BO) is an efficient method to optimize expensive black-box functions. It has been generalized to scenarios where objective function evaluations return stochastic binary feedback, such as success/failure in a given test, or preference between different parameter settings. In many real-world situations, the objective function can be evaluated in controlled `contexts' or `environments' that directly influence the observations. For example, one could directly alter the `difficulty' of the test that is used to evaluate a system's performance. With binary feedback, the context determines the information obtained from each observation.  For example, if the test is too easy/hard, the system will always succeed/fail, yielding uninformative binary outputs. Here we combine ideas from Bayesian active learning and optimization to efficiently choose the best context and optimization parameter on each iteration. We demonstrate the performance of our algorithm and illustrate how it can be used to tackle a concrete application in visual psychophysics: efficiently improving patients' vision via corrective lenses, using psychophysics measurements.

 \end{abstract}

\section{Introduction}
%Consider the case where we are trying to optimize an unkown black-box function 
%\begin{equation}
%	 \arg \min g(x)
%\end{equation}
%Standard BO ... 
%
%\begin{equation}
%	p(c=1|x) = \Phi(g(x)) 
%\end{equation}
%
%
%\begin{equation}
%	p(c=1|x,s) = \Phi(f(s,x)) 
%\end{equation}
%where 
%\begin{equation}
%	\arg \min_x f(s,x) = \arg \min_x g(s,x)
%\end{equation}

% Bayesian optimization (BO) is an efficient method to maximize the performance of expensive black-box functions. 

Bayesian optimization (BO) has been used in many applications to optimize systems whose performance is expensive to evaluate, such as a robot's walking motion \citep{Antonova2017}, deep brain stimulation \citep{Grado2018}, or machine learning model hyperparameters \citep{Snoek2012}. BO has also been generalized to situations where one can only obtain the performance of the system in the form of a binary output, corresponding, for example, to whether the system passes/fails a test \citep{Tesch2013}. Examples of binary Bayesian optimization (BBO) include testing whether a robot is able to successfully maneuver across a given environment, or the optimization of a model’s hyperparameters, by early-stopping the training of under-performing models while continuing others \citep{Zhang2019a}.
In these cases, the way the system is tested is critical in optimizing its performance. For example, if we ask a robot to move across an impossibly complex environment (e.g.~filled with obstacles and pot-holes), it will always fail, leaving us none the wiser as to how to improve its performance.

A similar problem arises in preferential Bayesian optimization (PBO), where the performance of a system is evaluated via a set of comparisons between different parameter settings \citep{Brochu2010b}. For example, PBO was used to improve the performance of hearing aid devices, by asking patients to evaluate which of two parameter settings they preferred, when listening to excerpts of music \citep{Nielsen2015a}. In this example, the context (i.e.~the audio stimuli) clearly plays a role in determining how much information is gained from subjects' responses: an audio stimulus that was overly simple (e.g. a single monotone pitch) or random (e.g. white noise) would result in non-informative responses from subjects, impeding progress in optimizing the hearing device for most daily settings.

Here we show how to extend previous BO algorithms to deal with these two related problems, in which the measured performance of a system takes the form of a binary, context-dependent, output. We then show how this problem can be efficiently solved by applying a simple heuristic, where BO is used to maximize the system's performance, while Bayesian active learning is used to choose a new context, such that the observed binary output (e.g. whether the system passed/failed a given test, or a comparison between two parameter settings) is maximally informative about the system's performance. We show that our resulting algorithm performs competitively on a range of benchmark problems. Further, we illustrate the relevance of our algorithm by applying it to a real-world problem, where the goal is to find the correct optical correction required to optimize a patient's vision, based on their binary responses in a psychophysics task. 

%Our work is a first towards the generalization of previous algorithms that were restricted to the continuous, noiseless feedback setting  

% \matty{Previous works looked at contextual BO in the restricted case where we have access to continuous, noiseless feedback} \citep{Toscano-Palmerin2018, Pearce2020, Weichert2020}. Specifically, they considered scenarios in which the objective can be directly measured, but its value depends on the context, and the goal is to maximize the expected value (across random context) or the worst-case value.

% To summarise, the contribution of our work is as follows:
% \begin{itemize}
% 	\item We generalize contextual Bayesian optimization to the case of binary outputs. 
% 	\item  We show that with binary outputs, setting the context correctly is particularly important, so as to avoid the cases where the test is too easy/hard, and thus, the observed output is almost always 0 or 1. 
% 	\item We show this problem can be efficiently solved using a simple heuristic, that combines aspects of BO and Bayesian active learning.
% \end{itemize}

\section{Problem statement}
Consider a system, whose performance can be quantified by a function, $g(\boldsymbol{x})$. For example, $g\br{\b x}$ could represent the visual acuity (VA) of a patient fitted with glasses with optical parameters, $\b x$. 
We are interested in solving the global optimization problem:
 \begin{equation}
\label{cBBO_problem}
\b x^*=\underset{\b x \in \mathcal{X}}{\arg \max } \ g\br{\b x},
\end{equation}
where  $\mathcal{X}$ is a bounded set denoted as the search space.

We consider the case where the performance of the system cannot be evaluated directly, but instead, can be assessed by observing whether it passes/fails a test in a given context, parameterized by $\b s$. For example, we could test whether a subject (fitted with glasses with optical parameters, $\b x$) can correctly identify a visual stimulus,  parameterized by  $\b s$, presented onscreen. The outcome of such a test is denoted by a binary variable, $c$, with success probability given by:
\begin{equation}
p\br{c=1|\b x, \b s}   = \Phi\br{f\br{\b s, \b x}}, \label{eq:response_prob}
\end{equation} 
where $\Phi$ is the normal cumulative distribution function, and  $f\br{\b s, \b x}$ is a function describing performance of the system with parameters $\b x$ and given a context $\b s$.

If we are to optimize the system based on observed binary outcomes, $c$, the probability that the system passes a test in a given context, $\b s$ (determined by $f\br{\b s, \b x}$) must be closely related to its overall performance, $g\br{\b x}$. For simplicity, we begin by assuming that the same configuration, $\b x^*$, maximizes performance in all contexts, such that:
\begin{equation}
\forall \boldsymbol{s} \in \mathcal{S}, \hspace{0.3cm}\underset{\boldsymbol{x} \in \mathcal{X}}{\arg \max } \ g\left(\boldsymbol{x}\right) =\underset{\boldsymbol{x} \in\mathcal{X}}{\arg \max } \ f\left(\boldsymbol{s, x}\right), 
\label{condition}
\end{equation}
where $\mathcal{S}$ is a bounded set denoted as the context space. In our previous example, this would imply that the optical correction that maximizes VA also maximizes the subject's performance in identifying a range of different visual stimuli, parameterized by $\b s$. This condition will be true for a broad range of situations in which $f\br{\b s, \b x}$ takes the form:
\begin{equation}
	f\br{\b s, \b x} = q\br{h\br{\b s} g \br{\b x} + b\br{\b s}}
\end{equation}
where $h\br{\b s}$ is non-negative for all $\b s$, and $q\br{\cdot }$ is a monotonically increasing one-dimensional function. Plugging this back into Eqn \ref{eq:response_prob}, we have:
\begin{equation}
	p\br{c=1|\b s, \b x} = \Phi\sqbr{ q\br{h\br{\b s} g\br{\b x} + b\br{\b s}}}.
\end{equation}

We can think of $b\br{\b s}$ and $h\br{\b s}$ as controlling the difficulty of the test. Without loss of generality, we can assume that  $g\br{\b x}\geq 0$ for all $\b x$. For large $h\br{\b s}$ the probability of success will saturate close to 1 for all $\b x$: i.e., the task is easy, so the system always passes the test. In contrast, when $h\br{\b s}$ is close to zero, the  system will perform close to baseline, with success probability $\Phi(q(b(\boldsymbol{s})))$. In both these extreme cases, little information is gained about the parameters, $\b x$, that optimize performance, since the success probability depends only weakly on $g\br{\b x}$. Our goal, therefore, is to find a sampling rule that selects $(\b s$, $\b x)$ on each trial so as to appropriately set the difficulty of the task, and efficiently optimize $g\br{\b x}$ in a limited number of trials.

\section{Contextual binary Bayesian optimization}

\subsection{Inference}
As is standard in BO, we will build a surrogate model of $f\br{\b s, \b x}$, by assuming a zero-mean Gaussian process (GP) prior, such that $f\sim GP\br{0, k\br{\cdot, \cdot}}$, where $k$ is a kernel function. When the observed outcome is binary, the posterior probability distribution over $f$ cannot be written analytically but must be approximated. While a number of different approximations schemes exist, here we use the Laplace approximation, since it is standard and simple to implement \citep{Rasmussen2006}. However, our approach could be applied to other approximation schemes such as expectation propagation \citep{Minka2001, Seeger2002}. 

Ideally, the kernel should reflect prior knowledge about the objective, such as the fact that the optimum does not depend on the context. This can be done, for example, by assuming a multiplicative structure of the function of the form $f(\boldsymbol{s},\boldsymbol{x}) = h(\boldsymbol{s})g(\boldsymbol{x})$, which in practice can be incorporated by using a kernel that decomposes as: $k(\boldsymbol{s},\boldsymbol{x}, \boldsymbol{s}', \boldsymbol{x}') = k_1(\boldsymbol{s},\boldsymbol{s}')k_2(\boldsymbol{x},\boldsymbol{x}')$. Moreover, $h$ should be constrained to be positive. In practice, we did not use this latter constraint in our experiments, as this would make the inference more difficult and it did not seem to impede performance.   
%\matty{You should say something here about the form of the kernel, which has to reflect our assumptions about the form of f(x,s). 
%Also, there is the problem (that we discussed), that h(x) needs to be constrained to be positive for constraint (3) to hold strictly.}

\subsection{Knowledge gradient acquisition rule}
Having inferred a Bayesian model of the objective function, $p\br{f|\mathscr{D}}$, the next step of any BO algorithm is to select new parameters, $\b x$ and $\b s$, to evaluate the objective. To do this, we extended the `knowledge gradient' (KG) acquisition rule, developed previously for BO with continuous outputs \citep{Frazier2009}, to contextual BBO. 

We first explain our binary KG algorithm for standard BBO (i.e with no contextual variable, $\b s$). As in standard KG, the reported solution of the optimization routine at iteration $t$ is assumed to be the maximum of the posterior mean of the GP after observing data $\mathscr{D}_t = \br{\{\b x_1,  c_1\}, \{\b x_2,  c_2\}, \dots \{\b x_t,  c_t\}}$, that is:
\begin{equation}
    \mu^*_{\mathscr{D}_t} = \underset{\b x \in \mathcal{X}}{\arg \max} \ \mathbb{E}\sqbr{g(\boldsymbol{x})|\mathscr{D}_t}. \label{maximu}
\end{equation}
The KG corresponds to the expected increase in $\mu^*$ if we are allowed one additional observation, $\{\boldsymbol{x}_{t+1}, c_{t+1}\}$:

\begin{eqnarray}
\mathrm{KG}(\boldsymbol{x}_{t+1})  &=& \sum_{c_{t+1}=0,1}p(c_{t+1} |\boldsymbol{x}_{t+1}, \mathscr{D}_t) \mu^*_{\mathscr{D}_t, \b x_{t+1}, c_{t+1}} \nonumber \\&&- \mu_{\mathscr{D}_t}^*
\end{eqnarray}
where $p(c_{t+1}=1| \boldsymbol{x}_{t+1}, \mathscr{D}_t) = \mathbb{E}_g[\Phi(g(\boldsymbol{x}_{t+1}))|\mathscr{D}_t]$, and  $ \mu^*_{\mathscr{D}_t, \b x_{t+1}, c_{t+1}}$ is the maximum of the posterior mean after observing $(\mathscr{D}_t, \{\boldsymbol{x}_{t+1}, c_{t+1}\})$. 
In supplementary \ref{app:KGgradient} we show that the gradient of KG, required to maximize it efficiently, can be computed when using the Laplace approximation (see supplementary \ref{app:Laplace}) to approximate the GP posterior.

To extend the above KG acquisition function to contextual BBO, we simply add the context, $\b s$, to the system parameters, $\b x$, and replace $g\br{\b x}$ with $f\br{\b s, \b x}$. In this case, the maximization in Eqn \ref{maximu} is performed only over the system parameters, $\b x$, given a fixed context, $\b s_0$ (see supplementary \ref{supp:max_KG}).

Computing the KG acquisition function is impractically slow for problems with moderately high dimensions, as its computation requires performing two nested optimizations: the optimization required for the Laplace approximation of the GP posterior (see supplementary \ref{supp:max_KG}), and the maximization in Eqn \ref{maximu}. Thus, we looked for an alternative, that could scale to larger problems.

\subsection{Sequential algorithm for choosing system parameters, $\b x$ and context, $\b s$}
In our set-up, the parameters, $\b x$, and $\b s$, play two very different roles. The system parameters $\b x$ determine the system's underlying performance, $g\br{\b x}$, which we want to maximize. The context, $\b s$, determines how much information we gain about the system's performance from binary observations, $c$. We therefore decided to choose $\b x$ and $\b s$ sequentially, using different heuristics: first, $\b x_{t+1}$ was chosen using binary BO, so as to optimize the system's performance; next, $\b s_{t+1}$ was chosen using Bayesian active learning,  to maximize the information we obtain from the binary observation, $c_{t+1}$.

\subsubsection{Choosing $\b x$ using Bayesian optimization}
We tested two different acquisition rules for selecting $\b x_{t+1}$: GP Upper Credible Bound (GP-UCB) \citep{Srinivas2010}, and Thompson Sampling (TS) \citep{Thompson1933}. 

Thompson sampling (TS) involves choosing $\boldsymbol{x}_{t+1}$ by sampling from the distribution $p(\boldsymbol{x}^* |\mathscr{D}_t),$
where $\b x^*$ is defined, for a given objective function, as  $\b x^* = \arg \max_{\b x} f\br{\b x, \b s}$ (note that, from Eqn \ref{condition},  $\b x^*$ is independent of $\b s$). In supplementary section \ref{sub:efficiently_sampling_GP} we describe how to efficiently sample from $p(\boldsymbol{x}^*|\mathscr{D}_t)$ in the context of GP classification and preference learning.
%TS shows competitive performance in standard BO  (see e.g. \citet{Basu2017}) as well as its generalizations to Preferential BO \citep{Gonzalez2017a, Sui2017a}. 
%Briefly, following \citet{Hernandez-Lobato2014} we first draw a sample from a finite-dimensional approximation to the posterior GP, then maximize this sample to find $\boldsymbol{x}_{t+1}$. To draw the sample from the posterior GP, we used the decoupled-bases approximation by \citet{Wilson2020a} jointly with the reduced-rank approximation described in \citet{Solin2020a}. 

Second, we implemented a generalization of UCB to binary outputs, where $\b x_{t+1}$ is chosen by maximising the following acquisition function:
\begin{equation}
\alpha_{\mathrm{UCB}}(\boldsymbol{x}) = \mathbb{E}(\Phi(f(\boldsymbol{x,\b s})) + \beta \sqrt{\mathbb{V}(\Phi(f(\boldsymbol{x, \b s}))},
\label{eq:binary_UCB}
\end{equation} 
where a constant $\beta$ (set to $\beta = \Phi^{-1}(0.95)$) sets the balance between exploitation (first term on right hand side) and exploration (second term on right hand side). See \citet{Fauvel2022} for details on how to analytically compute this acquisition function and its gradient.

\subsubsection{Choosing $\b s$ using Bayesian active learning}
After choosing the system parameters, $\boldsymbol{x}_{t+1}$, the next step is to select a context $\boldsymbol{s}_{t+1}$ such that the observation, $c_{t+1}$ will be maximally informative about the underlying function $f(\cdot, \boldsymbol{x}_{t+1})$. To do this, we used the Bayesian Active Learning by Disagreement (BALD) algorithm, developed by \citet{Houlsby2011}.

For a given $\boldsymbol{x}_{t+1}$, the context $\boldsymbol{s}_{n+1}$ is chosen so as to maximize the mutual information between the observation $c_{t+1}$ and the objective function $f\br{\cdot, \b x_{t+1}}$, $I(c_{t+1}, f | \boldsymbol{s}_{t+1}, \boldsymbol{x}_{t+1})$, which is given by: 
\begin{eqnarray}
I(c_{t+1}, f | \boldsymbol{s}_{t+1}, \boldsymbol{x}_{t+1}) &=&H(c_{t+1}| \boldsymbol{x}_{t+1}, \boldsymbol{s}_{n+1}) \\&&- \mathbb{E}_f[H(c_{t+1}| \boldsymbol{x}_{t+1}, \boldsymbol{s}_{n+1}, f)], \nonumber
\end{eqnarray}
where $H(c_{t+1}| \boldsymbol{x}_{t+1}, \boldsymbol{s}_{n+1})$ is the total entropy of $c_{t+1}$, and $\mathbb{E}_f[H(c_{t+1}| \boldsymbol{x}_{t+1}, \boldsymbol{s}_{n+1}, f)]$ is the conditional entropy of $c$, given the latent function $f\br{\cdot, \b x_{t+1}}$. (Note that all three terms are also conditioned on the observed data, $\mathscr{D}_t$,  which was removed from the equation for notational simplicity). This acquisition criterion can be efficiently approximated \citep{Houlsby2011}.

\subsection{Preferential Bayesian optimization}
We next considered how to extend our method to preferential optimization (PBO). In standard PBO \citet{Brochu2010b, Gonzalez2017a,Dewancker2018}, 
i.e.~with no contextual variable, $\b s$, the result of an evaluation, $c$, depends on the relative value of the objective function, $g$, with two different parameter settings, $\b x$ and $\b x'$, according to: 
\begin{equation}
    P(c=1| g, \b x, \b x') = \Phi\br{ g\br{\b x}- g\br{\b x'}}.
\end{equation}
The parameters to be compared, $\b x$ and $\b x'$, are called a duel. To extend this to the contextual case, we assume that the comparison depends on both the system parameters, $\b x$, and context, $\b s$, according to:
\begin{equation}
P(c=1| f,\boldsymbol{s}, \b x, \b x')= \Phi\left(f\left(\boldsymbol{s, x}\right) - f\left(\boldsymbol{s, x'}\right)\right),
\end{equation}
where $f\br{\b s, \b x}$ is defined as before, such that, for all $\b s\in \mathcal S$, $ \arg \max_{\b x}  g\left(\boldsymbol{x}\right) =\arg \max_{\b x }  f\left(\boldsymbol{s, x}\right)$ (Eqn \ref{condition}). 

The generalization of the KG to this scenario is straightforward. Likewise, our sequential acquisition rule can be generalized to this scenario by selecting the duel $(\boldsymbol{x},\boldsymbol{x'})$ using PBO, and then choosing the context of the duel $\boldsymbol{s}$ using Bayesian active learning.

In our experiments, we selected system parameters using either: (i) the KernelSelfSparring (KSS) algorithm \citep{Sui2017}, which is a simple extension of TS, described above, or (ii) the Maximally Uncertain Challenge (MUC) acquisition rule, described in \citet{Fauvel2022}. Contextual parameters, $\b s$, were then selected using Bayesian active learning by disagreement (BALD; see previous section).

\section{Experiments with synthetic test functions}

To evaluate the performance of our method, we ran optimization experiments on a set of 34 functions from a widely used virtual library for optimization experiments \citep{Surjanovic}. The functions in this library exhibit a diversity of behaviors that occur in real-life optimization problems. 

For each objective function, we inferred the hyperparameters for three different kernels (squared exponential, Matérn 3/2 and Matérn 5/2) using maximum likelihood estimation with 1000 randomly chosen samples. We then determined for each function the kernel that best described the function by measuring the root-mean-squared error on 3000 points. The benchmark functions are listed in supplementary \ref{app:benchmarks}. In all cases, we used the Laplace approximation to approximate the posterior over the objective, $p\br{f|\mathscr{D}}$. 

We introduced a scalar context variable $s\in [0,1]$, so that for a test function $g$ the response of a system query is 1 with probability $P( s, \boldsymbol{x}) = \Phi(sg(\boldsymbol{x}))$ and 0 otherwise.

To take the context variable into account when building the GP surrogate model, we used the following kernel: for a base kernel $k$ determined using the aforementioned procedure, we used: 
$k_s((\boldsymbol{x},s), (\boldsymbol{x'},\b s')) = ss'k(\boldsymbol{x}, \boldsymbol{x}')$.

 To compare the different algorithms, we used the stratified analysis method proposed by  \citet{Dewancker2016}. Briefly, for each benchmark function, we performed pairwise comparisons between acquisition functions using the Mann-Whitney U test at $\alpha = 5\times 10^{-4}$ significance on the best value found at the end of the optimization sequence. This determines a partial ranking based on the number of wins.

Ties are then broken by running the same procedure, but based on the Area Under Curve, which is related to the speed at which the algorithm reaches the optimum. This generates a new partial ranking, based on which a Borda score \citep{Dwork2001} is attributed to each acquisition function (the Borda score of a candidate is the number of candidates with a lower rank). Then, rankings from different benchmarks are aggregated by summing the Borda scores to establish a global ranking. This can be seen as a weighted vote from each benchmark function.
 
\subsection{Binary Bayesian optimization in adaptive contexts}
We then ran optimizations on each function for 40 different random number generator seeds, on 60 iterations. The initial number of random samples was set to 5.

 We termed the sequential acquisition rules in the binary feedback scenario TS with active learning by disagreement (TS-ALD) and GP-UCB with active learning by disagreement (UCB-ALD). We termed the generalization of KSS and MUC to the contextual scenario KSS-ALD and MUC-ALD.

We compared the different acquisition rules: the contextual binary knowledge gradient (cBKG), UCB-ALD and TS-ALD, with the following controls: 
\begin{itemize}
\item Fully random, in which $(s, \boldsymbol{x})$ was chosen at random at each iteration,
\item $\boldsymbol{x}$ chosen using TS and $s$ chosen randomly,
\item $\boldsymbol{x}$ chosen using UCB and $s$ chosen randomly,
\item $(s, \boldsymbol{x})$ selected using BALD.
\end{itemize}

\begin{figure*}[tb]
\centering
\includegraphics[width =\textwidth]{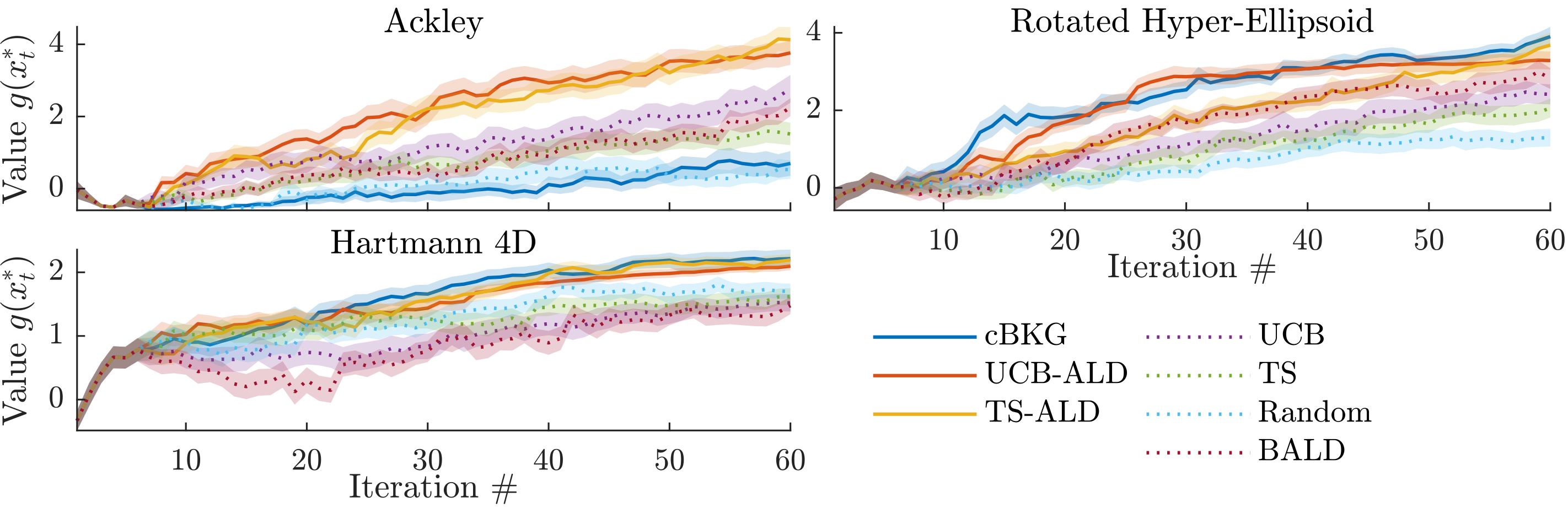}
\caption[Preference learning with adaptive contexts]{Results across three different objective functions and different methods in binary BO with adaptive contexts.  Solid lines correspond to the mean value of the inferred maximum across the 40 repetitions. Shaded areas correspond to standard error on the mean. Either cBKG, UCB-ALD, or TS-ALD outperform the other acquisition rules, however, there is a lot of variability from one experiment to another. The cBKG acquisition rule tends to be less robust than the sequential heuristics.}
\label{fig:cBBO}
\end{figure*}

To avoid saturation effects when transforming the benchmark functions through the non-linearity, we scaled the functions so that they have mean 0 and variance 1. Examples of regret curves are presented in figure \ref{fig:cBBO}, and the results of the stratified analysis are summarized in table \ref{tab:cBBO_benchmarks}. More detailed results showing pairwise comparisons between acquisition functions are presented in table \ref{fig:cBBO_matrices}.

UCB-ALD and TS-ALD show superior performance compared to cBKG (\ref{tab:cBBO_benchmarks}), despite the fact that, compared to cBKG, where $\boldsymbol{x}$ and $\boldsymbol{s}$ are jointly selected, these sequential decision strategies induce an adaptivity gap \citep{Jiang2019b}. 
 Here, we used the Laplace approximation, given that Expectation Propagation (EP) is known to improve performance compared to the Laplace approximation and that the gradient of cBKG with EP seems intractable, this suggests that TS-ALD and UCB-ALD are better acquisition functions in general. The three heuristics outperformed the different controls.

\begin{table}[!htp]\centering
\caption[Comparison of acquisition rules on benchmarks]{Comparison of acquisition functions on benchmarks. UCB-ALD largely outperforms the other rules. Importantly, the sequential acquisition rules (TS-ALD and UCB-ALD) outperform cBKG.}
\begin{tabular*}{\columnwidth}{@{\extracolsep{\fill}} ccc}
Acquisition rule & Rank & Borda score \\ 
\toprule
UCB-ALD & 1 & 144 \\ 
TS-ALD    & 2 & 83 \\ 
cBKG                            & 3 & 59 \\ 
UCB (random context)                     & 4 & 46 \\ 
BALD                   & 5 & 35 \\ 
TS (random context)                        & 6 & 19 \\ 
Random                                                    & 7 & 10 \\ 
\bottomrule
\end{tabular*}
\label{tab:cBBO_benchmarks}
\end{table}

\subsection{Preferential Bayesian optimization in adaptive contexts }
We then evaluated the performance of our sequential acquisition rules in their generalization to the case of preferential judgments: KSS-ALD and MUC-ALD. Given the limited performance of BKG in the previous experiments and the fact that the preferential version of Knowledge Gradient is impractically slow, we did not evaluate this acquisition function.

We then ran optimizations on each test function for 40 different random number generator seeds, on 60 iterations. 
We compared  KSS-ALD and MUC-ALD to the following controls: 
\begin{itemize}
\item Fully random, in which $(s, \boldsymbol{x}_1, \boldsymbol{x}_2)$ was chosen at random at each iteration,
\item $\boldsymbol{x}$ chosen using KSS and $s$ chosen randomly,
\item $\boldsymbol{x}$ chosen using MUC and $s$ chosen randomly,
\item $(s, \boldsymbol{x}_1, \boldsymbol{x}_2)$ selected using BALD.
\end{itemize}

\begin{figure*}[tb]
\centering
\includegraphics[width =\textwidth]{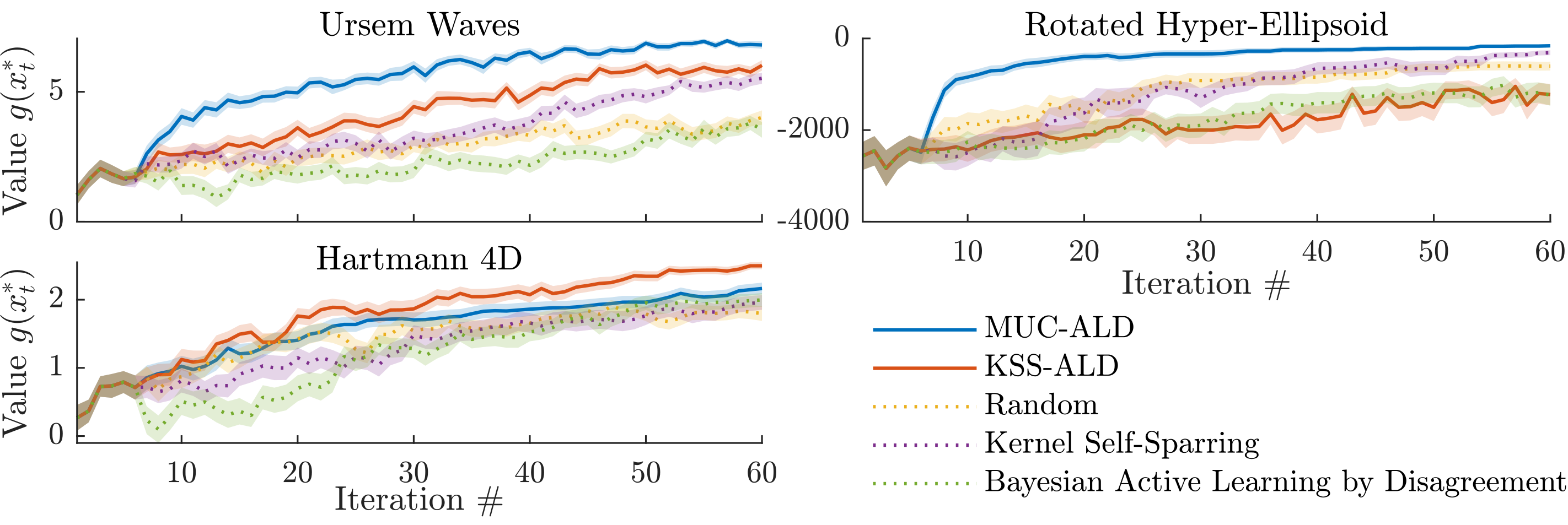}
\caption[Preference learning with adaptive contexts]{Results across three different objective functions in preferential BO with adaptive contexts.  Solid lines correspond to the mean value of the inferred maximum across the 40 repetitions. Shaded areas correspond to standard error on the mean. Sequential acquisition rules (KSS-ALD and MUC-ALD) consistently outperform the others.}
\label{fig:cPBO}
\end{figure*}

Examples of regret curves are shown in figure \ref{fig:cPBO}, and the results of the stratified analysis are summarized in figure \ref{tab:cPBO_benchmarks}. More detailed results showing pairwise comparisons between acquisition functions are presented in table \ref{fig:cPBO_matrices}.
Again, adaptive selection of the context leads to superior performance compared to controls.

\begin{table}[!htp]\centering
\caption[Comparison of acquisition rules on benchmarks]{Comparison of acquisition functions on benchmarks. The sequential acquisition rules outperform the controls. MUC-ALD is the most efficient acquisition function.}
\begin{tabular*}{\columnwidth}{@{\extracolsep{\fill}} ccc}
\toprule
MUC-ALD                                  & 1 & 99 \\ 
KSS-ALD                                  & 2 & 37 \\ 
Kernel Self-Sparring                     & 3 & 22 \\ 
BALD & 4 & 9 \\ 
Random                                   & 5 & 5\\ 
\bottomrule
\end{tabular*}
\label{tab:cPBO_benchmarks}
\end{table}

\section{Application: adaptive optimization in psychometric measurements}
To illustrate the relevance of our algorithm, we considered the problem of optimizing the parameters of a lens to improve a patient's vision based on their responses in a visual task.

Different types of visual defects typically result in different types of error (myopia results in `spherical refractive error', $S$; astigmatism results in `cylindrical refractive error', $C$). These errors induce image blur, but can be corrected by optimizing different optical parameters of a patient's lenses.  However, patients' visual performance are typically assessed indirectly, for example by testing their performance in a visual task (e.g. asking patients to identify a letter, presented onscreen). Clearly, the task used to test patients' visual performance will play a critical role in optimizing their vision. If the task is too easy/difficult (e.g. the letters presented onscreen are too small/large) then patients will always succeed/fail, and little information will be gained as to how to optimize their optics. In the following, we show our contextual BO algorithm can be used to address this problem. 

%the eye is too large compared to its focusing power, which induces spherical refractive error. In astigmatism, the cornea is ellipsoid instead of spherical, causing a cylindrical refractive error.
%Half of the world's population is expected to be myopic (short-sighted) by 2050, and automated measurement of refractive correction has become an important issue \citep{Holden2016}.
%To correct for these errors, an ophthalmologist alters two optical parameters of the corrective lenses, called the `vergence' and the `
%both the vergence of the optics (the parameter to optimize) and the size of the optotypes (standardized figures used to measure VA, such as the Snellen letters) in standard recognition tasks. If the letters are too big or too small, the responses will be uninformative about the optimal correction. The usual correction step is 0.25 diopters ($\delta$), but recent technologies, such as Essilors' Advanced Vision Accuracy (Essilor International, Charenton-le-Pont, France) allow for 0.01 $\delta$ precision. Moreover, many patients combine several types of refractive errors, such as myopia and astigmatism, which requires tuning several parameters of the lens at the same time. As a consequence, subjective refractive error measurements need to be adapted to be rapid and precise.

%We evaluated contextual binary BO in a simulation of a patient's subjective refractive error measurement. To do so, we first built a model of the response in a visual task where the patient has to identify an optotype displayed on a screen.

\subsection{Model of patients' responses}

A simple formula relates the spherico-cylindrical error (measured in diopters ($\delta$)) to image blur $\beta$ \citep{Raasch1995, Blendowske2015}:
\begin{equation}
\beta =  \sqrt{\frac{1}{2}(S^2+(S+C)^2)}
\end{equation}We are thus trying to find the parameters of the optics $\boldsymbol{x} = (S,C)$ that minimize $\beta$, or equivalently, minimize visual acuity (VA), defined as the log of the minimum angle of resolution (MAR, measured in minutes of arc).
VA can  be related to blur using the formula \citep{Blendowske2015}: $$\text{VA}(\boldsymbol{x}) - \text{VA}(\boldsymbol{x^\star}) = 1+\beta^2,$$
where $\boldsymbol{x^\star}$ corresponds to the best correction. A VA of 20/20 corresponds to logMAR = 0, so we considered that $\text{VA}(\boldsymbol{x^\star}) = 0$.

We consider a simple model of subjects performing an $n$-alternatives forced-choice visual task, such as identifying a letter on a screen. The size of the letter, $s$ (measured in log of the visual angle, in minutes of arc), determines the difficulty of this task. For a given optical correction $\boldsymbol{x}$, the probability of correct response is well described by a psychometric function (see e.g. \citet{Fulep2019} or \citet{Tokutake2011}): 
\begin{equation}
P(c = 1| s) = \gamma + (1-\gamma) \Phi(a(\boldsymbol{x})s + b(\boldsymbol{x})),
\end{equation}
where $\gamma$ is the chance probability of success,  $a$ and $b$ are the slope and intercept of the psychometric curve. $a$ is assumed to be strictly positive. Here, for simplicity and consistently with experimental evidence, we assumed that the slope of the psychometric function $a(\boldsymbol{x})$ remains essentially constant with varying blur \citep{Horner1985,Carkeet2001}.
%To improve a patient's vision, we would like to improve their visual acuity (VA). 

VA is usually defined as the value of $s$ at which $P(c = 1|s) = \frac{1+\gamma}{2}$, or equivalently the inflexion point of the psychometric curve: $VA(\boldsymbol{x}) = -\frac{b(\boldsymbol{x})}{a}$.

To apply our algorithm to this problem, we define a function: 
\begin{equation}
f(s, \boldsymbol{x}) = \Phi^{-1}[\gamma + (1-\gamma) \Phi(a s + b(\boldsymbol{x}))]
\label{eq:f}
\end{equation}
such that: $P(c = 1| s, \boldsymbol{x}) = \Phi(f(s, \boldsymbol{x}))$ (as in Eqn 2). The problem of improving VA  thus satisfies condition \ref{condition}, as for any $s$, the maximum of $f$ is the minimum of VA.

\subsection{Gaussian process model}

The use of GP classification for psychometric function estimation has been introduced by \citet{Gardner2015a, Song2017, Song2018} in the context of audiometry. We follow the same approach and build a GP model of the subject's responses in the task. 

For $\gamma$ large enough, we have $P(c = 1| s) \approx \Phi(a(\boldsymbol{x})s + b(\boldsymbol{x}))$. Since in practice, with a letter chart, $\gamma^{-1}= n=26$, we will make this simplifying assumption in building our surrogate model. 

We put a GP prior on $f$, with zero mean and kernel $k_\psi$ defined as:
\begin{equation}
k_\psi((s, \boldsymbol{x}), (s', \boldsymbol{x}')) = \theta ss' + k(\boldsymbol{x},\boldsymbol{x}')
\end{equation}
This kernel reflects the structure of the function $f$ (see equation \ref{eq:f}) with the assumption $\gamma\approx0$ (see supplementary \ref{supp:kernel_psychophysics}).

We used a squared exponential kernel as $k$. To minimize the effect of kernel hyperparameters, we simulated the response to 1000 random pairs $(s, \boldsymbol{x})$ and inferred the kernel hyperparameters using maximum likelihood estimation. The hyperparameters were then kept constant during the experiments.

\subsection{Results}
\begin{figure}[tb]
\centering
\includegraphics[width =\columnwidth]{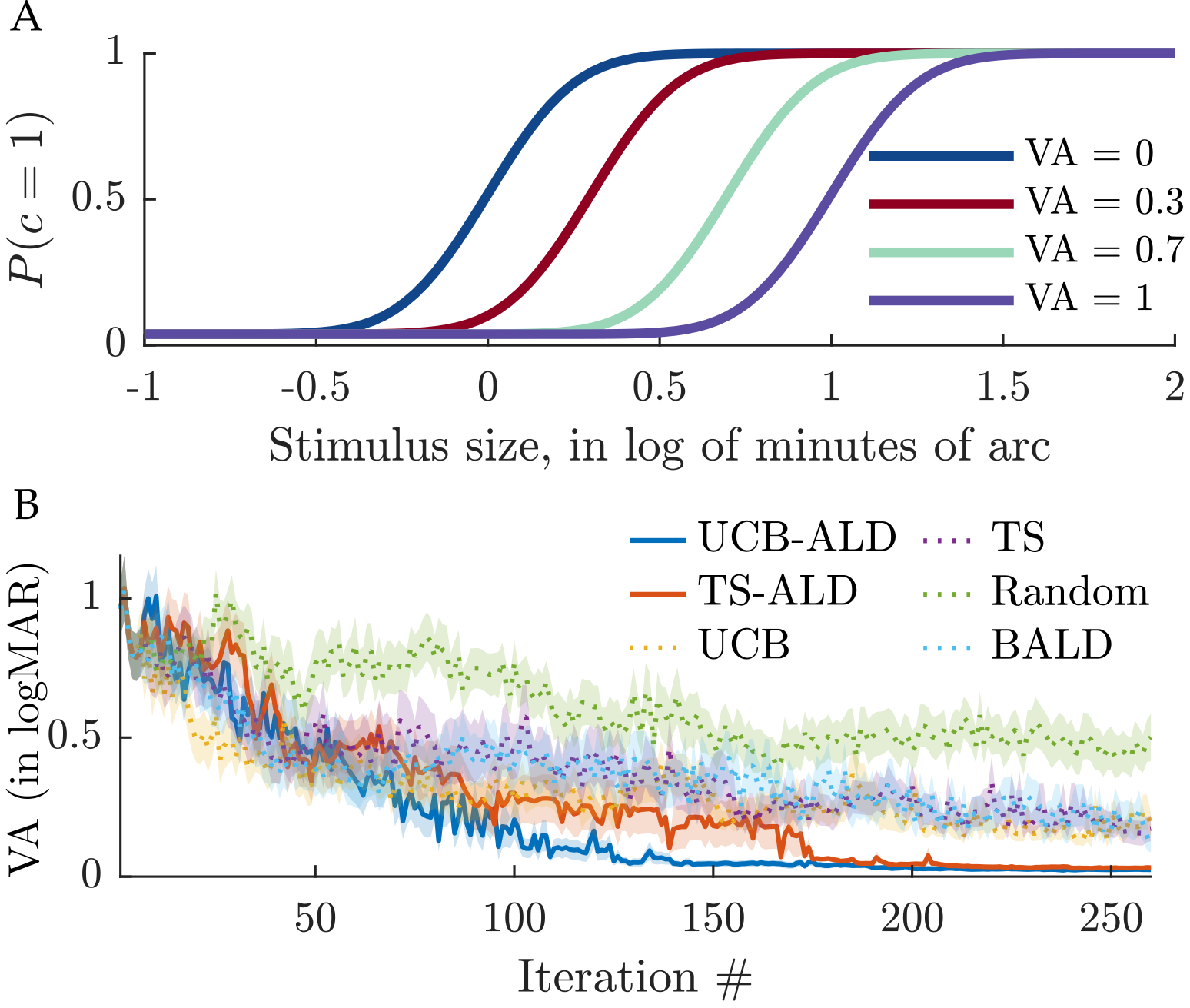}
\caption[Refractive error correction]{\textbf{A.} Psychometric curves for various refractive errors corresponding to different visual acuities, with a slope of the psychometric function set at  5.0 logMAR$^{-1}$. The baseline corresponds to the chance level in the n-alternatives forced-choice task (here, n = 26). VA of 20/20 corresponds to logMAR = 0.  \textbf{B.} Visual acuity with the inferred best parameters throughout the optimization. The lines correspond to the average over 60 repetitions, whereas the shaded areas correspond to standard error on the mean.  UCB-ALD  and TS-ALD both lead to faster convergence towards optimal correction (logMAR = 0) compared to controls where the context (TS, UCB, random)  or the correction (BALD, random) are not adaptively selected.}
\label{fig:VA}
\end{figure}

We repeated the simulated experiment 20 times for 260 iterations, for 8 different slopes values (evenly spaced between between 1.0 logMAR$^{-1}$ and  8.0 logMAR$^{-1}$). The search space was $\mathcal{X} = [-4 \delta,4\delta] \times [-4\delta,4\delta]$ and the context space $ \mathcal{S} = [-1, 2] $ logMAR. Regret curves in the experiments where the slope of the psychometric function was set at  5.0 logMAR$^{-1}$ are shown on figure \ref{fig:VA}, and the results of the stratified analysis are summarized in table \ref{tab:cBBO_exp}.   
 
 Both UCB-ALD and TS-ALD led to rapid and consistent improvement of the VA closed to its optimal value of logMAR = 0. UCB-ALD is the best performing algorith (see table \ref{tab:cBBO_exp}), with a mean VA at the end of the optimization sequence of $2.48 \times 10^{-2}$ logMAR in the case of a slope of  5.0 logMAR$^{-1}$(s.e.m = 8.22), \ref{fig:VA}). At the end of the optimization sequence, the an average spherical correction error of $2.74 \times 10^{-2}\delta$  (s.e.m. = $5.36 \times 10^{-2}$). Note that this is closed to the precision that can be achieved nowadays ($0.01\delta$). The average cylindrical correction error is  $1.53 \times 10^{-1}\delta$  (s.e.m. = $8.22 \times 10^{-2}$).   
 
 However, when using random sampling, the optimization consistently failed (figure \ref{fig:VA}).  Overall, the results are consistent with the one obtained on the synthetic benchmarks (table \ref{tab:cBBO_exp}).
 
 Here, we considered that the objective is a black box. However, for a specific application in a clinical setting, parameterizing the objective using domain knowledge would likely lead to faster and more robust improvement (see e.g. \citet{Cox2017}).

\begin{table}[!htp]\centering
\caption[Comparison of acquisition rules on benchmarks]{Comparison of acquisition functions on the refractive error correction experiment with 8 different slopes (evenly spaced between between 1.0 logMAR$^{-1}$ and  8.0 logMAR$^{-1}$). The two heuristics combining BO and active learning ( UCB-ALD and TS-ALD) outperform the controls where the context (TS, UCB, random)  or the correction (BALD, random) are not adaptively selected.}
\begin{tabular*}{\columnwidth}{@{\extracolsep{\fill}} ccc}
Acquisition rule & Rank & Borda score \\ 
\toprule
UCB-ALD & 1 & 35 \\ 
TS-ALD   & 2 & 26 \\ 
UCB (random context)                                      & 3 & 15 \\ 
BALD               & 4 & 8 \\ 
TS (random context)                        & 5 & 3 \\ 
Random                                                    & 6 & 0 \\ 
\bottomrule
\end{tabular*}
\label{tab:cBBO_exp}
\end{table}

\section{Discussion}

\subsection{Summary of contributions}
%In this paper, we introduced a new framework for BO:  BO in adaptive contexts, in which the experimenter can choose the context where each function evaluation is performed. 
In this paper, we introduced a new framework for binary and preferential BO:  BO in adaptive contexts, in which the experimenter can choose the context where each function evaluation is performed. We proposed an acquisition function to jointly select the optimization variable and the context and showed how to generalize existing acquisition functions to this framework by combining them with Bayesian active learning. We show on synthetic benchmarks that the proposed algorithms outperform controls in which the context is not adaptively selected, in both BOs with binary or preference feedbacks.
Finally, we showcase our framework on a concrete problem: tuning the parameters of the lens for patients with refractive errors.

\subsection{Related work}

%Our work is related to BO with environmental variables \citep{Klein2017, McLeod2017}: in this scenario, some environments correspond to cheap but unreliable evaluations, and others to accurate but expensive evaluations. Our framework differs as we do not consider the cost of the queries, and we do not know a priori how informative a given context is. 

In the bandit setting, contextual GP bandit \citep{Krause2011} extends the GP bandit framework to scenarios in which at each round corresponds a given random context, and actions are taken knowing this context. The goal is to identify the best arm averaged over all possible contexts. This framework has been generalized to dueling bandits \citep{Dudik2015}. Our work differs as we consider the problem of both selecting a context and action at each iteration. 

%In Offline Contextual BO \citep{Char2019}, each context corresponds to a different objective and the goal is to solve these different optimization problems jointly. 

The methods most related to our work are BO with expensive integrands \citep{Toscano-Palmerin2018}, conditional BO \citep{Pearce2020}, and contextual policy search \citep{Pinsler2019}, where the goal is to maximize the expected performance conditionally on a random context variable. During the optimization process, this context variable is controlled by the experimenter. 
In BO with expensive integrands \citep{Toscano-Palmerin2018}, the goal is to maximize a function averaged over a set of random conditions, when function evaluation is noise-free or with i.i.d. Gaussian noise. The authors used a generalization of the knowledge gradient. \citet{Pearce2020} further generalized by considering the case where function evaluations are performed in parallel and batch of inputs are selected at each iteration.  
Our work is a generalization to the binary and preferential feedback scenarios, which implies new challenges. Indeed, even when the objective function is the same across different contexts, the amount of information we get from each evaluation varies with the context in a way that is a priori unknown. This problem arises in general when the amount of noise in function evaluation varies with the context, but we only considered the case of probit likelihoods for simplicity. We also introduced sequential acquisition rules that allow the generalization of acquisition rules used in BO. We did not consider whether sequential acquisition also improves performance in the case of Gaussian likelihood, and this should be the topic of future work. In this paper, we considered the case where the optimum of the function $f$ is the same across all contexts. In general, the maximum of $f(\boldsymbol{s}, \cdot)$ may depend on $\boldsymbol{s}$. In that case,  several criteria may be considered. Either we may want to maximize the averaged performance of the system across contexts, or we may want to maximize the worst-case performance of the device. We let the study of these generalizations for future work.

%Potential future developments also include the generalization of our method to batches, where several points are selected and evaluated simultaneously. 

%\citet{Thatte2017} applied the Predictive Entropy Search acquisition rule \citep{Hernandez-Lobato2014} to PBO. The Predictive Entropy Search with Preferences (PES-P) samples comparisons that optimally reduce the uncertainty in the distribution of the objective function maximum with the least number of queries and it could in theory be applied in the contextual scenario. However, this acquisition function requires several approximations and is slow to compute.

   %The former scenario corresponds to the ConBO framework \citep{Pearce2020}, whereas the second corresponds to Max-Min BO \citep{Weichert2020}. Our approach is a generalization of these frameworks as we consider cases in which the context directly determines the amount of information about $f$ at the sampled points.  

%The generalization of the ConBO framework to the binary case is given by:

\subsection{Code availability}
Matlab implementation available at \url{https://disclose_upon_release}.

\bibliography{Thesis_citations}
\bibliographystyle{icml2021}

\begin{appendices}


% Supplementary material: To improve readability, you must use a single-column format for the supplementary material.
\onecolumn

\aistatstitlesupp{Contextual Bayesian optimization with binary outputs: \\
Supplementary Materials}

\setcounter{equation}{0}
\setcounter{figure}{0}
\setcounter{table}{0}
\setcounter{page}{1}
\setcounter{section}{0}
\makeatletter
\renewcommand{\theequation}{S\arabic{equation}}
\renewcommand{\thefigure}{S\arabic{figure}}
\renewcommand{\thetable}{S\arabic{table}}
\renewcommand{\bibnumfmt}[1]{[S#1]}
\renewcommand{\citenumfont}[1]{S#1}

\section{The Laplace approximation for Gaussian process classification}
\label{app:Laplace}
The following description of Laplace approximation for Gaussian process classification is given for completeness. It is required in order to understand the computation of the gradient of the Binary Knowledge Gradient. We follow the reasoning presented in \citet{Rasmussen2006} as well as \citet{Bishop2006}, to which we refer the reader for further details.

In Gaussian process classification, we assume that observations $\boldsymbol{c}$ at points $\boldsymbol{X} = [\boldsymbol{x}_1, \cdots, \boldsymbol{x}_t]$ are Bernoulli random variables with parameters $ \mu_c(\boldsymbol{X}) = \pi(f(\boldsymbol{X}))$, where $f$ is a latent function, and and $\pi$ is an inverse link function. We choose the convention where $c$ is either 0 or 1.

The predictive distribution at point $\boldsymbol{x}$ is given by:

\begin{equation}
p(c=1|\boldsymbol{x}, \mathscr{D}) = \int p(c=1| f(\boldsymbol{x}))p(f(\boldsymbol{x})|\mathscr{D})df(\boldsymbol{x}).
\label{pc}
\end{equation}
Since this integral is analytically intractable, we approximate $p(f(\boldsymbol{x})|\mathscr{D})$ with a Gaussian distribution. Indeed, for a a random normally distributed random variable $z$: 
\begin{equation}
    \int \Phi(z)\mathcal{N}(z|\mu, \sigma^2)dz = \Phi\left(\frac{\mu}{\sqrt{1+\sigma^2}}\right) .
\end{equation}

A convenient way to do so is to note that: 
\begin{equation}
p(f(\boldsymbol{x})|\mathscr{D}) = \int p(f(\boldsymbol{x})|\boldsymbol{f})p(\boldsymbol{f}|\mathscr{D})d\boldsymbol{f},
\label{predictive}
\end{equation}
where $\boldsymbol{f}$ is the vector of latent values at training points $X$.
From the formula for Gaussian process posteriors, we have: 
\begin{equation}
p(f(\boldsymbol{x})|\boldsymbol{f}) = \mathcal{N}\left(f(\boldsymbol{x})|\boldsymbol{k}^\top  \boldsymbol{K}^{-1}\boldsymbol{f}, k(\boldsymbol{x}, \boldsymbol{x}) - \boldsymbol{k}^\top  \boldsymbol{K}^{-1}\boldsymbol{k}\right),
%\label{GP_condition}
\end{equation}where $\boldsymbol{K} = k(\boldsymbol{X},\boldsymbol{X})$ and $\boldsymbol{k} = k(\boldsymbol{X},\boldsymbol{x})$.

The second term in the integral $p(\boldsymbol{f}|\mathscr{D})$ is the posterior distribution of the latent value function at training points. By approximating it with a Gaussian distribution, then we could compute the integral in \ref{predictive}, which would give us a Gaussian approximation for $p(f(\boldsymbol{x})|\mathscr{D}) $. This approximation would in turn allow us to compute \ref{pc}.

\subsection{Principle of the Laplace approximation}

We start by finding a Gaussian approximation of $p(\boldsymbol{f}|\mathscr{D})$. In general, for a random variable $\boldsymbol{z}$ whose probability density function is $p(\boldsymbol{z}) = \frac{f(\boldsymbol{z})}{Z}$, we can use a second a second order Taylor expansion around the mode $z_0$ of the distribution (where the gradient vanishes) so that: 
\begin{equation}
\ln f(\boldsymbol{z}) \sim \ln f(\boldsymbol{z_0})  - \frac{1}{2}(\boldsymbol{z}-\boldsymbol{z_0})^\top \boldsymbol{H}(\boldsymbol{z}-\boldsymbol{z_0}),
\end{equation} 
where $\boldsymbol{H}$ is the negative of the Hessian of $f$ at $\boldsymbol{z_0}$.

Taking the exponential and computing the appropriate normalization coefficients, we have: 
\begin{equation}
p(\boldsymbol{z}) \sim \frac{|\boldsymbol{H}|^\frac{1}{2}}{(2\pi)^\frac{D}{2}}\exp\left(-\frac{1}{2}(\boldsymbol{z}-\boldsymbol{z_0})^\top \boldsymbol{H}(\boldsymbol{z}-\boldsymbol{z_0})\right) = \mathcal{N}(\boldsymbol{z}|\boldsymbol{z_0}, \boldsymbol{H}^{-1})
\end{equation}

This method of approximating a probability density function with a Gaussian probability density function is called the Laplace approximation. 

\subsection{Gaussian approximation of the posterior}

In order to find a Gaussian approximation of $p(\boldsymbol{f}|\mathscr{D})$, we thus need to compute its mode and its Hessian.
By using Bayes' rule, we have: 

\begin{equation}
\ln p(\boldsymbol{f}|\mathscr{D}) = \ln p(\boldsymbol{f}) + \ln p(\mathscr{D}|\boldsymbol{f}).
\end{equation} 
The prior term is:
\begin{equation}
\ln p(\boldsymbol{f}) = -\frac{1}{2}\boldsymbol{f}^\top \boldsymbol{K}^{-1}\boldsymbol{f}-\frac{t}{2}\ln(2\pi)-\frac{1}{2}\ln|K|.
\end{equation}
The likelihood term is: 
\begin{equation}
\begin{aligned}
\ln p(\mathscr{D}|\boldsymbol{f})  &= \ln \left(\prod_{i=1}^t \pi(f_i)^{c_i}(1-\pi(f_i))^{(1-c_i)} \right) \\
&= \sum_{i=1}^t \ln \left(\pi(f_i)^{c_i}(1-\pi(f_i))^{(1-c_i)}\right).
\end{aligned}
\end{equation}
So the gradient of the the log-posterior is: 

\begin{equation}
\nabla_{\boldsymbol{f}} \ln p(\boldsymbol{f}|\mathscr{D}) = \nabla_{\boldsymbol{f}} \ln p(\mathscr{D}|\boldsymbol{f}) - \boldsymbol{K}^{-1}\boldsymbol{f},
\end{equation}
whereas the Hessian is: 
\begin{equation}
\nabla_{\boldsymbol{f}}^2 \ln p(\boldsymbol{f}|\mathscr{D}) = \nabla_{\boldsymbol{f}}^2 \ln p(\mathscr{D}|\boldsymbol{f}) - \boldsymbol{K}^{-1}, 
\end{equation}
where $\nabla^2$ refers to the Hessian matrix.

We introduce $W = -\nabla_{\boldsymbol{f}}^2 \ln p(\mathscr{D}|\boldsymbol{f})$, which is a diagonal matrix since conditionally on $\boldsymbol{f}$, observations are independent. 

\begin{equation}
\frac{\partial \ln p(c_i|f_i)}{\partial f_i} = \frac{\pi'(f_i)(c_i-\pi(f_i))}{\pi(f_i)(1-\pi(f_i))}.\end{equation}

In the case where the link function is the cumulative normal distribution: 
\begin{equation}
\frac{\partial \ln p(c_i|f_i)}{\partial f_i} = \frac{(2c_i-1)\phi(f_i)}{\Phi((2c_i-1)f_i)},\end{equation}
and 
\begin{equation}
\frac{\partial^2 \ln p(c_i|f_i)}{\partial f_i^2} = -\frac{\phi(f_i)^2}{\Phi((2c_i-1)f_i)^2} - \frac{(2c_i-1)f_i\phi(f_i)}{\Phi((2c_i-1)f_i)}.\end{equation}

The mode $\boldsymbol{f}_0$ satisfies the condition  $\nabla_{\boldsymbol{f}} \ln p(\boldsymbol{f}|\mathscr{D}) = 0$, so $\boldsymbol{f}_0 = \boldsymbol{K}\nabla_{\boldsymbol{f}} \ln p(\mathscr{D}|\boldsymbol{f})$.
The mode is usually found using the Newton-Raphson method. 

The approximate posterior distribution is :
\begin{equation}
p(\boldsymbol{f}|\mathscr{D}) = \mathcal{N}(\boldsymbol{f}_0, (\boldsymbol{K}^{-1}+\boldsymbol{W})^{-1}).
\label{Approximation}
\end{equation}

\subsection{Approximate predictive distribution}

By combining equation \label{GP_condition} with equation  \ref{Approximation}, we get: 
\begin{equation}
\mathbb{E}_f[f(\boldsymbol{x})|\mathscr{D}] =  \boldsymbol{k}^\top \boldsymbol{K}^{-1}\boldsymbol{f}_0,
\end{equation}
and: 
\begin{equation}
\mathbb{V}_f[f(\boldsymbol{x})|\mathscr{D}] = k(\boldsymbol{x}, \boldsymbol{x}) - \boldsymbol{k}^\top (\boldsymbol{K}+\boldsymbol{W}^{-1})^{-1}\boldsymbol{k}.
\end{equation}

\clearpage
\section{Gradient of the knowledge gradient}
\label{app:KGgradient}
The knowledge gradient is defined as: \begin{equation}
\mathrm{KG}(\boldsymbol{x})  = \mathbb{E}_{c\sim p(c|\mathscr{D})}(\mu_{n+1}^\star - \mu_{n}^\star | \mathscr{D}, \boldsymbol{x}_{n+1} = \boldsymbol{x}),
\end{equation}
where $\mu_{n+1}^\star$is the maximum of the posterior mean after observing $(\mathscr{D}, (\boldsymbol{x}, c))$, and  $\mu_{n}^\star$ is the maximum of the posterior mean after observing $\mathscr{D} = (\boldsymbol{X}, c_{1\cdots, n})$.

With binary outputs, the knowledge gradient can be expressed as: 
\begin{equation}
\mathrm{KG}(\boldsymbol{x})  = \mu_c(\boldsymbol{x})(\mu_{1, n+1}^\star - \mu_{n}^\star ) + (1-\mu_c(\boldsymbol{x}))(\mu_{0, n+1}^\star - \mu_{n}^\star ),
\end{equation}

where $\mu_c(\boldsymbol{x}) = \mathbb{E}_f[\Phi(f(\boldsymbol{x}))|\mathscr{D}]$, and  $\mu_{1,n+1}^\star$ (resp. $\mu_{0,n+1}^\star$)  is the maximum of the posterior mean after observing $(\mathscr{D}, (\boldsymbol{x}, 1))$ (resp.  $(\mathscr{D}, (\boldsymbol{x}, 0))$). That is: 

\begin{equation}
\mu_{c,n+1}^\star= \underset{\boldsymbol{x'} \in \mathcal{X}}{\max}\  \mathbb{E}_f[f(\boldsymbol{x}')|\mathscr{D}\cup(\boldsymbol{x}, c)].
\end{equation} 
The gradient of the knowledge-gradient is given by:
\begin{equation}
\nabla\mathrm{KG}(\boldsymbol{x}) = \nabla\mu_c(\boldsymbol{x})(\mu_{1,n+1}^\star - \mu_{0,n+1}^\star)+\mu_c(\boldsymbol{x})\nabla\mu_{1,n+1}^\star  + (1-\mu_c(\boldsymbol{x}))\nabla\mu_{0,n+1}^\star .
\end{equation}

To compute the  gradients of  $\mu_{1,n+1}^\star $ and $\mu_{0,n+1}^\star$ , inspired by \citet{Wu2016b}, we use the envelope theorem \citep{Milgrom2002}, which states that, under sufficient regularity conditions, the gradient with respect to $ \boldsymbol{x}$ of a maximum of a collection of functions of $\boldsymbol{x}$ is given simply by first finding the maximum $\boldsymbol{x}^\star$  in this collection, and then differentiating this single function with respect to $\boldsymbol{x}$, keeping $\boldsymbol{x}^\star$ fixed. 
Here, we have an infinite collection of functions  $\mathbb{E}_f[f(\boldsymbol{x}')|\mathscr{D}\cup(\boldsymbol{x}, c)]$ indexed by $\boldsymbol{x'}$. 

\begin{theorem}{Corollary 4 of the envelope theorem in \citet{Milgrom2002}}
Let $X$ denote the choice set and $t$ be a parameter in a $[0,1]$ (the theorem generalizes to normed vector spaces).
Let $f: X \times[0,1] \rightarrow \mathbb{R}$ be an objective function parameterized by $t$. We define: 
\begin{equation}
V(t) =  \underset{\boldsymbol{x} \in \mathcal{X}}{\sup}f(\boldsymbol{x}, t),\end{equation}
\begin{equation}
X^\star(t) = \{x\in \mathcal{X}, f(\boldsymbol{x},t) = V(t)\}.
\end{equation}

Suppose that X is a nonempty compact space, $f(\boldsymbol{x}, t)$ is upper semicontinuous in $\boldsymbol{x}$, and $\frac{\partial f}{\partial t}(\boldsymbol{x}, t)$ is continuous in $(\boldsymbol{x}, t)$. Then:
\begin{equation}
\forall t \in [0,1), V'(t+) =  \underset{\boldsymbol{x} \in \mathcal{X}^\star(t)}{\max}\frac{\partial f}{\partial t}(\boldsymbol{x}, t),\end{equation}
\begin{equation}
\forall t \in (0,1], V'(t-) =  \underset{\boldsymbol{x} \in \mathcal{X}^\star(t)}{\min}\frac{\partial f}{\partial t}(\boldsymbol{x}, t).\end{equation}
V is differentiable at any $t\in (0,1) $ if and only if $\left\{\frac{\partial f}{\partial t}(\boldsymbol{x}, t)|\boldsymbol{x}\in X^\star(t)\right\}$ is a singleton, and in that case $ \forall \boldsymbol{x}\in X^\star(t), V'(t) = \frac{\partial f}{\partial t}
(\boldsymbol{x}, \boldsymbol{t})$.
\end{theorem}

As a consequence, by writing $\boldsymbol{x}^\star = \underset{\boldsymbol{x'} \in \mathcal{X}}{\arg\max}\  \mathbb{E}_f[f(\boldsymbol{x}')|\mathscr{D}\cup(\boldsymbol{x}, c)]$, we get:
\begin{equation}
\nabla_{x}\mu_{c,n+1}^\star  = \nabla_{\boldsymbol{x}} \mathbb{E}_f[f(\boldsymbol{x}^\star)|\mathscr{D}\cup(\boldsymbol{x}, c)].\end{equation}
From the Laplace approximation, we have: 
\begin{equation}
\mathbb{E}_f[f(\boldsymbol{x}^\star)|\mathscr{D}\cup(\boldsymbol{x}, c)] = \boldsymbol{k}^\top \nabla_{\boldsymbol{y}}\log p(\boldsymbol{c}|\boldsymbol{y}),\end{equation}
were $\boldsymbol{y}$ corresponds to the inferred latent values of the training data, $\boldsymbol{c} = c_{1\cdots, n+1}$, and $\boldsymbol{k} = k((\boldsymbol{X},\boldsymbol{x}), \boldsymbol{x}^\star)$.

So that:

\begin{equation}
\begin{aligned}
\nabla_{\boldsymbol{x}}\mathbb{E}[f(\boldsymbol{x}^\star)|\mathscr{D}\cup(\boldsymbol{x},  c)] &= (\nabla_{\boldsymbol{x}}\boldsymbol{k}^\top )\nabla_{\boldsymbol{y}}\log p(\boldsymbol{c}|\boldsymbol{y}) +\boldsymbol{k}^\top \nabla_{\boldsymbol{x}}\nabla_{\boldsymbol{y}}\log p(\boldsymbol{c}|\boldsymbol{y}) \\
&= (\nabla_{\boldsymbol{x}}\boldsymbol{k}^\top )\nabla_{\boldsymbol{y}}\log p(\boldsymbol{c}|\boldsymbol{y}) +\boldsymbol{k}^\top (\nabla^2_{\boldsymbol{y}} \log p(\boldsymbol{c}|\boldsymbol{y}))\nabla_{\boldsymbol{x}}\boldsymbol{y}\\
& = (\nabla_{\boldsymbol{x}}\boldsymbol{k}^\top )\nabla_{\boldsymbol{y}}\log p(\boldsymbol{c}|\boldsymbol{y}) -\boldsymbol{k}^\top  \boldsymbol{W} \nabla_{\boldsymbol{x}}\boldsymbol{y},
\end{aligned}
\end{equation}
where $\boldsymbol{W} =-\nabla^2_{\boldsymbol{y}} \log p(\boldsymbol{c}|\boldsymbol{y})$.
 
From the Laplace approximation,  we have $\boldsymbol{y} = \boldsymbol{K}\nabla_{\boldsymbol{y}}\log p(\boldsymbol{c}|\boldsymbol{y})$, where $\boldsymbol{K} = k((\boldsymbol{X}, \boldsymbol{x}), (\boldsymbol{X}, \boldsymbol{x}))$.

By differentiating this self-consistent equation,

\begin{equation}
\begin{aligned}
\nabla_{\boldsymbol{x}}\boldsymbol{y} &= (\nabla_{\boldsymbol{x}} \boldsymbol{K})\nabla_{\boldsymbol{y}} \log p(\boldsymbol{c}|\boldsymbol{y})+\boldsymbol{K}(\nabla^2_{\boldsymbol{y}} \log p(\boldsymbol{c}|\boldsymbol{y}))\nabla_{\boldsymbol{x}}\boldsymbol{y} \\
&= (\nabla_{\boldsymbol{x}} \boldsymbol{K})\nabla_{\boldsymbol{y}} \log p(\boldsymbol{c}|\boldsymbol{y})-\boldsymbol{K}\boldsymbol{W}\nabla_{\boldsymbol{x}}\boldsymbol{y}.
\end{aligned}
\end{equation}

By rearranging: 
\begin{equation}
\nabla_{\boldsymbol{x}}\boldsymbol{y} = (\boldsymbol{I}+\boldsymbol{K}\boldsymbol{W})^{-1}\nabla_{\boldsymbol{x}}\boldsymbol{K}\nabla_{\boldsymbol{y}}\log (\boldsymbol{c}|\boldsymbol{y}).
\end{equation}

\section{Contextual binary knowledge gradient}
\label{supp:max_KG}

 With binary outputs and contexts, the knowledge gradient can be expressed as:
\begin{equation}
\label{KG_binary_context}
\mathrm{KG}(\boldsymbol{s}, \boldsymbol{x})  = \mu_c(\boldsymbol{s}, \boldsymbol{x})\mu_{t+1,1}^\star + (1-\mu_c(\boldsymbol{s}, \boldsymbol{x}))\mu_{t+1,0}^\star - \mu_{t}^\star,
\end{equation}
where $\mu_c(\boldsymbol{s}, \boldsymbol{x}) = P(c=1| \boldsymbol{s}, \boldsymbol{x}, \mathscr{D}_t) = \mathbb{E}_f[\Phi(f(\boldsymbol{s}, \boldsymbol{x}))|\mathscr{D}_t]$  is the predictive class distribution, and  $\mu_{t+1,1}^\star$ (resp. $\mu_{t+1,0}^\star$)  is the maximum of the posterior mean for an arbitrary context $\boldsymbol{s}_0$ after observing $(\mathscr{D}_t \cup  (\boldsymbol{x}, 1))$ (resp.  $(\mathscr{D}_t \cup (\boldsymbol{x}, 0))$). That is: 
 \begin{equation}
\mu_{t+1,c}^\star= \underset{\boldsymbol{x'} \in \mathcal{X}}{\max}\  \mathbb{E}_f[f(\boldsymbol{s}_0, \boldsymbol{x}')|\mathscr{D}_t\cup(\boldsymbol{s}_0, \boldsymbol{x}, c)].
\end{equation}Note that the term $\mu_{t}^\star$ in Eqn \ref{KG_binary_context} can be neglected, since it does not depend on $(\boldsymbol{s}, \boldsymbol{x}) $.

Importantly, if the kernel reflects the assumption that the maximum of the objective $f$ does not depend on the context  $\boldsymbol{s}$, then $\mu_{t+1,c}^\star$ can be computed for an arbitrary value of $\boldsymbol{s}_0$. Indeed, following the Laplace approximation (supplementary \ref{app:Laplace}), with latent values $\boldsymbol{f}_0$:

\begin{equation}
\mathbb{E}_f[f(\boldsymbol{s}, \boldsymbol{x})|\mathscr{D}] =  \boldsymbol{k}((\boldsymbol{s},\boldsymbol{x}), (S,X))^\top \boldsymbol{K}^{-1}\boldsymbol{f}_0,
\end{equation}
where $(S,X)$ corresponds to training points. Consider for example the case where $f(\boldsymbol{s}, \boldsymbol{x}) = h(\boldsymbol{s})g(\boldsymbol{x})$, the kernel reflecting this structure is of the form $k((\boldsymbol{s}, \boldsymbol{x}), (\boldsymbol{s}', \boldsymbol{x}')) = k_h(\boldsymbol{s}, \boldsymbol{s}')k_g(\boldsymbol{x}, \boldsymbol{x}')$. In that case, the posterior mean factorizes and it is clear that  we can take the maximum with respect to $\boldsymbol{x}$ independently of $\boldsymbol{s}$.

\clearpage
\section{Supplementary results}

\begin{figure}[h]
\centering
\includegraphics[width =\textwidth]{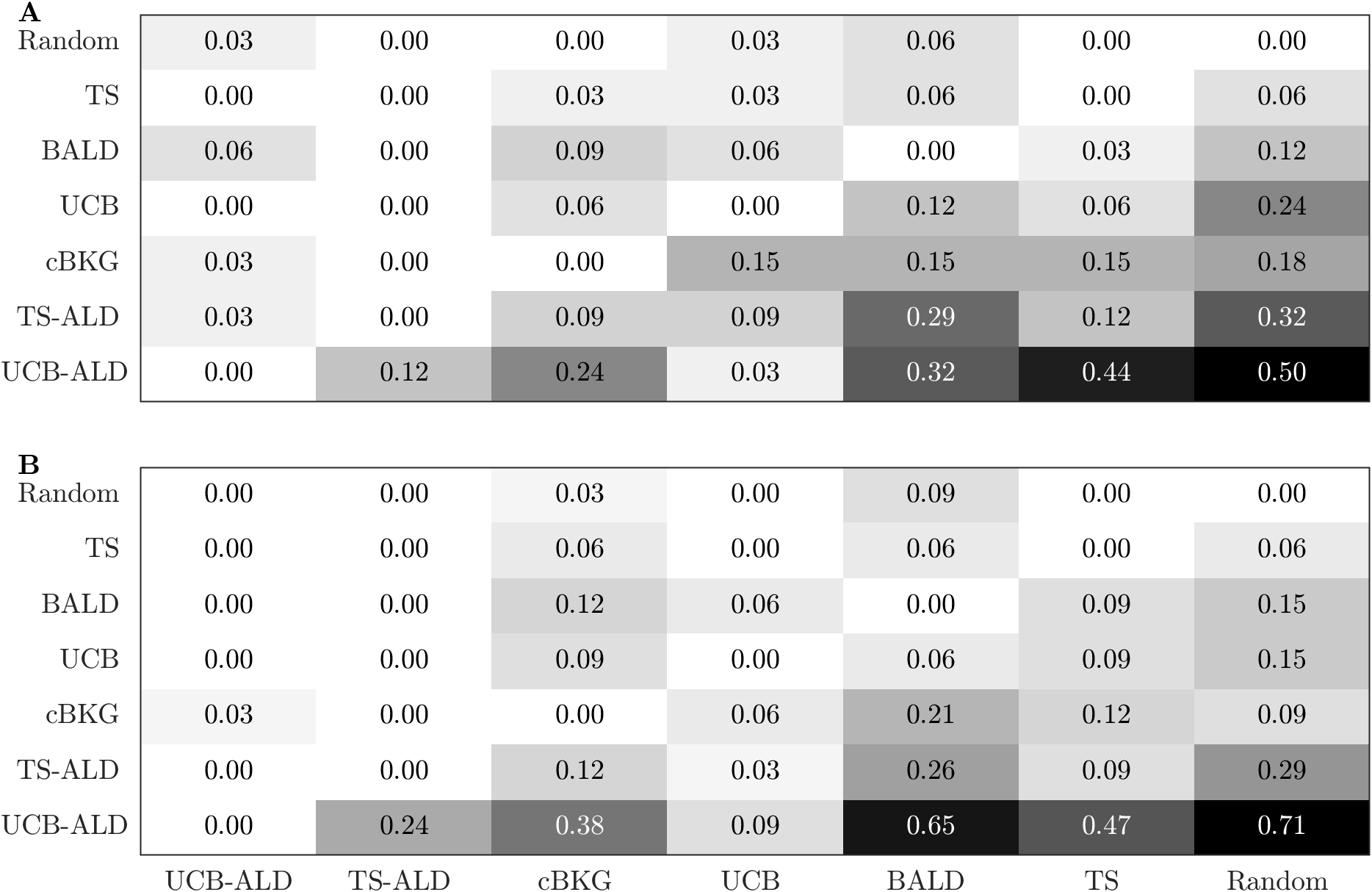}
\caption{Detailed results of performance comparison between acquisition functions in contextual Binary Bayesian optimization. Each entry (i,j) corresponds to the fraction of benchmarks functions for which i beats j according to the Mann-Whitney U test at $\alpha = 5\times 10^{-4}$ significance based either on the best value found (\textbf{A}) or the Area Under the Curve (\textbf{B}).}
\label{fig:cBBO_matrices}
\end{figure}

\begin{figure}[h]
\centering
\includegraphics[width =\textwidth]{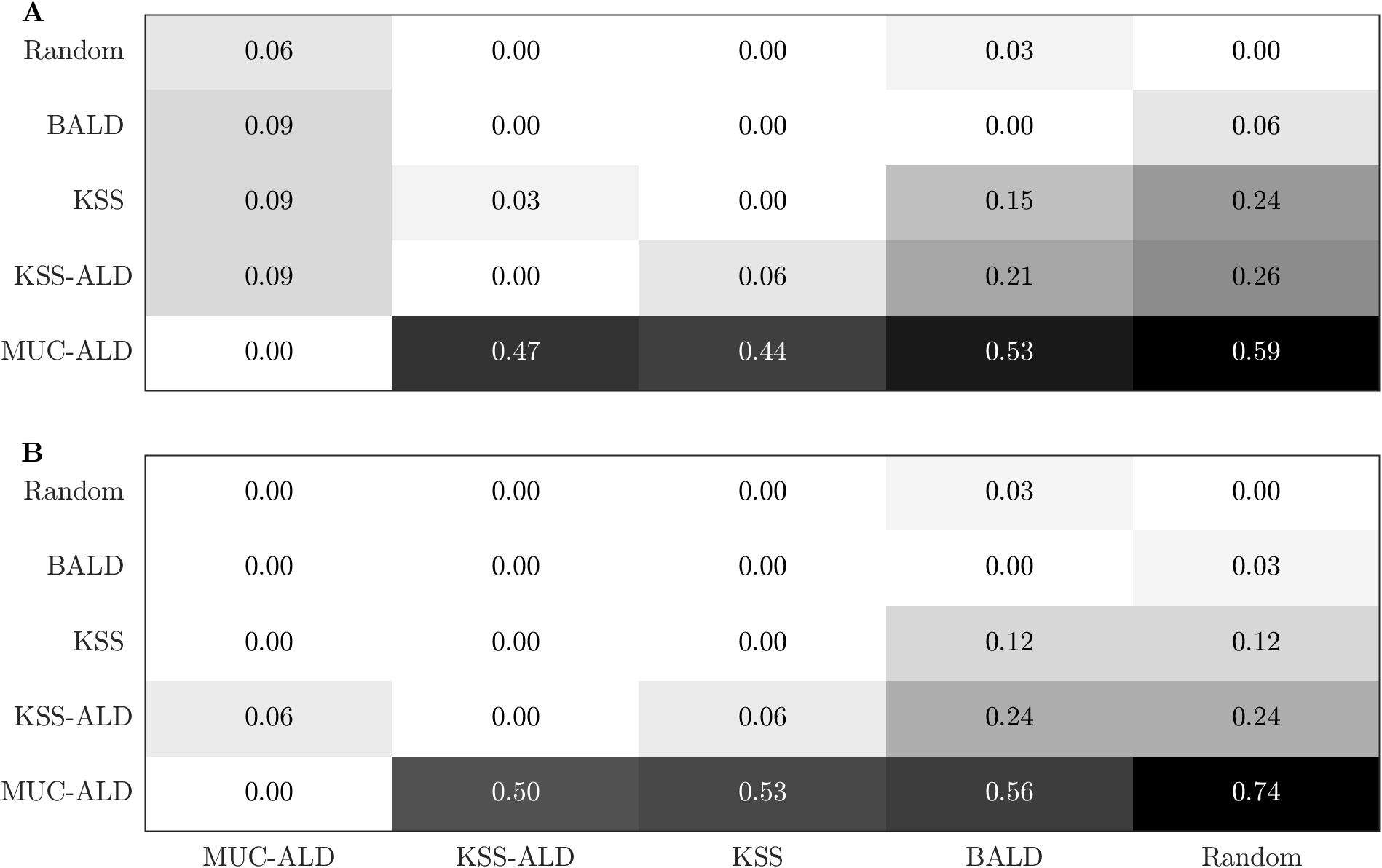}
\caption{Detailed results of performance comparison between acquisition functions in contextual Preferential Bayesian optimization. Each entry (i,j) corresponds to the fraction of benchmarks functions for which i beats j according to the Mann-Whitney U test at $\alpha = 5\times 10^{-4}$ significance based either on the best value found (\textbf{A}) or the Area Under the Curve (\textbf{B}).}
\label{fig:cPBO_matrices}
\end{figure}

\clearpage
\section{Kernel derivation for the psychophysics experiment}
\label{supp:kernel_psychophysics}

For lens parameters, $\boldsymbol{x}$, the response of the patient is determined by:
\begin{equation}
P(c = 1| s, \boldsymbol{x}) = \gamma + (1-\gamma) \Phi(a(\boldsymbol{x})s + b(\boldsymbol{x})).
\end{equation}
For $\gamma$ large enough, we have $P(c = 1| s) \approx \Phi(a(\boldsymbol{x})s + b(\boldsymbol{x}))$. Since in practice, with a letter chart, $\gamma^{-1}= n=26$, we will make this simplifying assumption in building our surrogate model. 

We put a GP prior on $f$, with zero mean and kernel $k_\psi$ defined as:
\begin{equation}
k_\psi((s, \boldsymbol{x}), (s', \boldsymbol{x}')) = \theta ss' + k(\boldsymbol{x},\boldsymbol{x}')
\end{equation}This kernel reflects the structure of the function $f$ (see equation \ref{eq:f}) with the assumption $\gamma\approx0$. Indeed, we have:
\begin{equation}
\begin{aligned}
\text{Cov}(f(s,\boldsymbol{x}), f(s',\boldsymbol{x}')) &= \text{Cov}(a(\boldsymbol{x})s+b(\boldsymbol{x}), a(\boldsymbol{x}')s'+b(\boldsymbol{x}')) \\ 
&=  ss'\text{Cov}(a(\boldsymbol{x})s+b(\boldsymbol{x}), a(\boldsymbol{x}')s'+b(\boldsymbol{x}')) \\
&+ (s+s')\text{Cov}(a(\boldsymbol{x}), b(\boldsymbol{x}'))+ \text{Cov}(b(\boldsymbol{x}), b(\boldsymbol{x}'))
\end{aligned}
\end{equation}
We put GP priors on functions $a$ and $b$, with kernels $k_1$ and $k_2$ respectively. since $a(\boldsymbol{x})$ and $b(\boldsymbol{x})$ are a priori independent conditionally on $\boldsymbol{x}$, the second term on the right-hand side vanishes, so that:
\begin{equation}
\begin{aligned}
\text{Cov}(f(s,\boldsymbol{x}), f(s',\boldsymbol{x}')) = ss'k_1(\boldsymbol{x},\boldsymbol{x}') + k_2(\boldsymbol{x},\boldsymbol{x}')
\end{aligned}
\end{equation}
Since we assumed the slope to be constant at value $\sqrt{\theta}$, this further simplifies to: 
\begin{equation}
\begin{aligned}
\text{Cov}(f(s,\boldsymbol{x}), f(s',\boldsymbol{x}')) = \theta ss' + k(\boldsymbol{x},\boldsymbol{x}')  
\end{aligned}
\end{equation}

 \section{Visual acuity as a function of the correction parameters}
 
 \begin{figure}[h]
\centering
\includegraphics{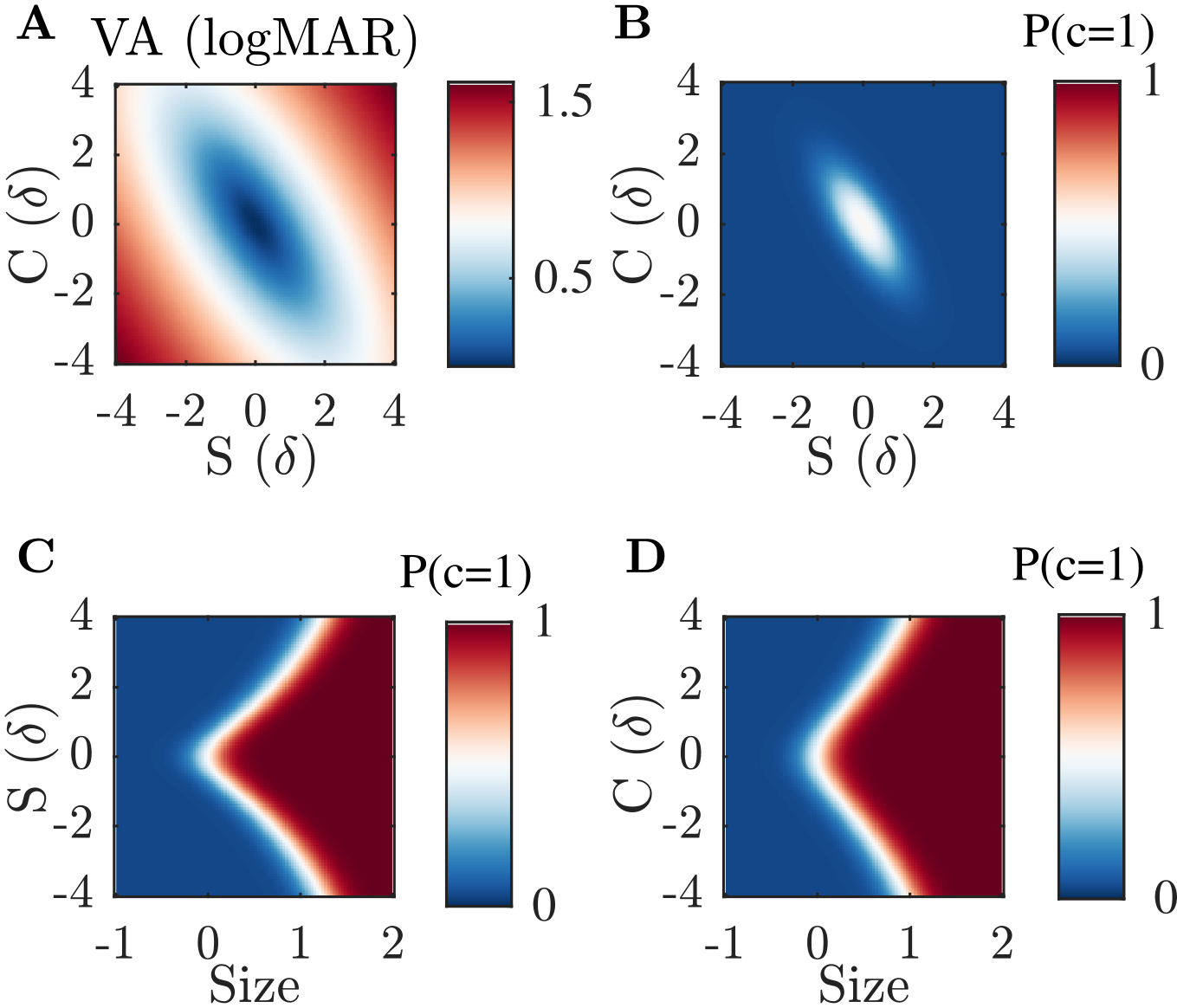}
\caption[Visual acuity model]{\textbf{A.} Visual acuity as a function of the correction parameters S (spherical correction) and C (cylindrical correction) in diopters ($\delta$).  The slope of the psychometric curve is kept fixed at 5.0 logMAR$^{-1}$. \textbf{B.} Probability of correct response in a 26-alternatives forced choice task for a stimulus of visual angle 1'. In blue regions, the subject will always perform at chance level, independently of the correction parameters, so responses will be uninformative.}
\label{fig:loss}
\end{figure}

\clearpage

\section{Sampling from GP classification models}
\label{sub:efficiently_sampling_GP}

Acquisition rules based on Thompson sampling rely on samples from the posterior distribution over the maximum: 
\begin{equation}
p\left(\boldsymbol{x}^\star \mid \mathscr{D}\right)=p\left(f\left(\boldsymbol{x}^\star\right)=\max _{\boldsymbol{x} \in \mathcal{X}} f(\boldsymbol{x}) \mid \mathscr{D}\right)
\end{equation}
\citet{Hernandez-Lobato2014} proposed the following sampling scheme: draw a sample from the posterior distribution $p\left(f | \mathscr{D}\right)$, then return the maximum of the sample. One could iteratively construct the sample $f$ while it is being optimized but, as noted by \citet{Hernandez-Lobato2014}, this would have a cost $\mathcal{O}\left(m^3\right)$, where $m$ is the number of evaluations of the function necessary to find the maximum. Although this is doable in practice,  \citet{Hernandez-Lobato2014}  suggested a more efficient procedure by sampling a finite-dimensional approximation to $f$, based on a finite-dimensional approximation to the kernel $k\left(\boldsymbol{x},\boldsymbol{x'}\right) \sim \phi(\boldsymbol{x})^\top \phi\left(\boldsymbol{x'}\right)$ \citep{Lazaro-Gredilla2010}. In GP classification and preference learning, this approximate sampling cannot be directly applied. In the following section, we will explain how to apply existing approximate sampling methods to the case of GP classification models.   

\subsection{Kernel approximation}
The sampling methods mentioned above consists in approximating a stationary kernel $k$ by means of the inner product of features $\phi$ such that: $k(\boldsymbol{x},\boldsymbol{x'}) \sim \phi(\boldsymbol{x})^\top \phi(\boldsymbol{x}')$.  
Recently, a method was proposed by \cite{Solin2020a}, which aims at making the approximation as good as possible for a given rank (see \citet{Riutort-Mayol2020} for details about the practical implementation). 
In this method, the kernel is approximated using a series expansion in terms of eigenfunctions of the Laplace operator on a rectangular domain $\Omega = [-L_1,L_1] \times \cdots \times [-L_d,L_d]$ (the search space is usually rectangular in Bayesian optimization).

 In preference learning, a specific difficulty arises. Indeed, the base kernel used to model the value function may be shift-invariant, the preference kernel, however, is not in general. 
This inexact hypothesis introduced in the sampling algorithm, leads to samples that are not consistent with the anti-symmetric property of a preference function, i.e. $f(\boldsymbol{x},\boldsymbol{x}') = -f(\boldsymbol{x}',\boldsymbol{x})$,  (see  e.g.  figure 4.1 in  \citet{Gonzalez2017a} where this inexact stationarity hypothesis is introduced). 

However, assume that we have a finite dimensional approximation to the base kernel $k(\boldsymbol{x},\boldsymbol{x}') \sim \phi(\boldsymbol{x})^\top \phi(\boldsymbol{x}')$, it is easy to see that we can approximate the preference kernel by $k_{\text{pref}}((\boldsymbol{x}_i,\boldsymbol{x}_j),(\boldsymbol{x}_k, \boldsymbol{x}_l)) \sim \phi_{\text{pref}}(\boldsymbol{x}_i, \boldsymbol{x}_j)^\top \phi_{\text{pref}}(\boldsymbol{x}_k, \boldsymbol{x}_l) $, with:
\begin{equation}
\label{eq:preference_features}
\phi_{\text{pref}}(\boldsymbol{x}_i, \boldsymbol{x}_j)= \phi(\boldsymbol{x}_i) - \phi(\boldsymbol{x}_j)
\end{equation} 
By construction, the corresponding sample is anti-symmetric.

\subsection{Weight-space approximation with non-Gaussian likelihoods}

The most widely used method for approximate sampling from GP in Bayesian optimization is the weight-space approximation. Assume that we have a finite-dimensional approximation to the kernel $k\left(\boldsymbol{x},\boldsymbol{x'}\right) \sim \phi(\boldsymbol{x})^\top \phi\left(\boldsymbol{x'}\right)$. The features $\phi(\boldsymbol{x})$ can be used to approximate the Gaussian process posterior with a Bayesian linear model: $f(\boldsymbol{x})\sim \phi(\boldsymbol{x})^\top \omega$, where \citep{Lazaro-Gredilla2010}:

\begin{equation}
\omega \sim \mathcal{N}\left(\left(\Phi^\top \Phi + \sigma^2I\right)^{-1}\Phi^\top \boldsymbol{y}, \left(\Phi^\top \Phi + \sigma^2I\right)^{-1}\sigma^2\right) 
\label{eq:sampling_w}
\end{equation} In the case of non-Gaussian likelihoods, , naively replacing $\boldsymbol{y}$ in \ref{eq:sampling_w} by the latent values inferred by the Laplace approximation (see \ref{app:Laplace}) or Expectation Propagation would not take into account the correlated heteroscedastic noise on the latent function values at training points. 
To the extent of our knowledge, the process of weight-space approximate sampling has not been rigorously introduced for latent GP models. Here, we suggest to use a sampling process in two steps. First, samples $\boldsymbol{y}$ are drawn from the posterior distribution over the latent variables at training points: $\mathcal{N}(\boldsymbol{\mu}, \boldsymbol{\Sigma})$, then $\boldsymbol{\omega}$ is sampled from $\boldsymbol{\omega} \sim \mathcal{N}\left(\left(\Phi^\top \Phi + \sigma^2I\right)^{-1}\Phi^\top \boldsymbol{y}, \left(\Phi^\top \Phi + \sigma^2I\right)^{-1}\sigma^2\right) $, where $\sigma$ is a small constant used for regularization. 

To see why this sampling scheme is correct, note that:
$p(f(\boldsymbol{x})|\mathscr{D}) = \int p(f(\boldsymbol{x})|X, \boldsymbol{y})p(\boldsymbol{y}|\mathscr{D})d\boldsymbol{y}$. So given latent values sampled from $p(\boldsymbol{y}|\mathscr{D})$, an approximate sample $\widetilde{f}$ can be drawn from $p(f(\boldsymbol{x})|X, \boldsymbol{y})$ using the method of \citet{Hernandez-Lobato2014}. 

 However, the degeneracy, i.e., low-rankness of the GP approximation, causes the estimate to grow over-confident when the number of observed points exceeds the degrees of freedom of the approximation. This results in ill-behaved approximations, and, in particular, underestimated variance, in regions far away from the data points. This phenomenon is known as variance starvation \citep{Wang2018a, Mutny2018, Calandriello2019a}.

\subsection{Decoupled-bases approximate sampling}

Recently \citet{Wilson2020a} proposed an efficient way to sample from GP posteriors that avoids variance starvation. The original sampling method was devised for exact GP with Gaussian noise and sparse GP, where the GP is computed based on a set of inducing points that explain the data, however, it can easily be generalized to non-Gaussian likelihood with a latent function \citep{Wilson2020b}. 

Briefly, this method is based on a corollary of Matheron's rule. For a Gaussian process $f \sim \mathcal{G} \mathcal{P}\left(0, k\right)$, the latent process conditioned on latent values  $(X, \boldsymbol{y})$  admits, in distribution, the representation: 
\begin{equation}
\begin{aligned}
\underbrace{\left(f \mid \boldsymbol{y}\right)\left(\cdot\right)}_{\scriptstyle \text {posterior}} \stackrel{\mathrm{d}}{=} \underset{\scriptstyle \text {prior}}{\underbrace{f\left(\cdot\right)}}+\underset{\scriptstyle \text {update}}{\underbrace{k\left(\cdot, \boldsymbol{x}\right) \mathbf{K}^{-1}\left(\boldsymbol{y}-f(X)\right)}}
\end{aligned}
\end{equation}

This corollary defines an approximation to the Gaussian process conditioned on $(X, \boldsymbol{y})$, where the stationary prior is approximated with a Bayesian linear model (weight-space prior), and the approximate posterior is obtained by adding an exact update (function-space update): 

\begin{equation}
\label{eq:Decoupled_bases}
\underbrace{\left(f \mid \boldsymbol{y}\right)\left(\cdot\right)}_{\scriptstyle \text{posterior}} \stackrel{\mathrm{d}}{\approx} \underbrace{\sum_{i=1}^{\ell} \omega_{i} \phi_{i}\left(\cdot\right)}_{_{\scriptstyle \text{weight-space prior}}}+\underbrace{\sum_{j=1}^{m} v_{j} k\left(\cdot, \boldsymbol{x}_{j}\right)}_{\scriptstyle \text{function space update}},
\end{equation}

 where $\boldsymbol{v}=\mathbf{K}^{-1}\left(\boldsymbol{y}-\boldsymbol{\Phi} \boldsymbol{\omega}\right)$, and $\boldsymbol{\omega}$ is sampled from $\mathcal{N}(\boldsymbol{0}, I)$. This method is termed decoupled-bases decomposition of the GP.
 
To sample from the posterior latent function, we thus sample   $\boldsymbol{\omega}$  from $\mathcal{N}(\boldsymbol{0}, I)$ and compute the corresponding weight-space prior, then sample $\boldsymbol{y}$ from $\mathcal{N}\left(\mu_f\left(X\right), \Sigma_f\left(X,X\right)\right)$ and compute the corresponding function-space update.

\clearpage
\section{Benchmarks}
\label{app:benchmarks}
 
\centering
\begin{longtable}{lccc}
Name & D & Kernel & Space \\ 
\toprule
\endhead
Ackley & 2 & Matérn 3/2 & $[-32.768,32.768]^2$ \\ 
Beale & 2 & SE-ARD & $[-4.5,4.5]^2$ \\ 
Bohachevsky & 2 & SE-ARD & $[-100,100]^2$ \\ 
Three-Hump Camel & 2 & Matérn 5/2 & $[-5,5]^2$ \\ 
Six-Hump Camel & 2 & SE-ARD & $[-3,3]\times [-2,2]$ \\ 
Colville & 4 & Matérn 5/2 & $[-10,10]^4$ \\ 
Cross-in-Tray & 2 & Matérn 5/2 & $[-10,10]^2$ \\ 
Dixon-Price & 2 & Matérn 5/2 & $[-5,5]^2$ \\ 
Drop-Wave & 2 & Matérn 3/2 & $[-5.12,5.12]^2$ \\ 
Eggholder & 2 & SE-ARD & $[-512,512]^2$ \\ 
Forrester et al (2008) & 1 & SE-ARD & $[0,1]$ \\ 
Goldstein-Price & 2 & SE-ARD & $[-2,2]^2$ \\ 
Griewank & 2 & SE-ARD & $[-600,600]^2$ \\ 
Gramacy and Lee (2012) & 1 & SE-ARD & $[0.5,2.5]$ \\ 
Hartmann 3-D & 3 & SE-ARD & $[0,1]^3$ \\ 
Hartmann 4D & 4 & SE-ARD & $[0,1]^4$ \\ 
Hartmann 6D & 6 & SE-ARD & $[0,1]^6$ \\ 
Holder & 2 & SE-ARD & $[-10,10]^2$ \\ 
Langer & 2 & Matérn 3/2 & $[0,10]^2$ \\ 
Levy & 2 & SE-ARD & $[-10,10]^2$ \\ 
Levy N.13 & 2 & Matérn 5/2 & $[-10,10]^2$ \\ 
Perm 0,d,$\beta$ & 2 & SE-ARD & $[-2,2]^2$ \\ 
Perm d,$\beta$ & 2 & SE-ARD & $[-2,2]^2$ \\ 
Powell & 4 & SE-ARD & $[-4,5]^4$ \\ 
Rosenbrock & 2 & SE-ARD & $[-2.048,2.048]^2$ \\ 
Rotated Hyper-Ellipsoid & 2 & Matérn 3/2 & $[-65.536,65.536]^2$ \\ 
Schaffer n4 & 2 & Matérn 3/2 & $[-100,100]^2$ \\ 
Schwefel & 2 & SE-ARD & $[-500,500]^2$ \\ 
Shekel & 4 & SE-ARD & $[0,10]^4$ \\ 
Schubert & 2 & Matérn 3/2 & $[0,10]^2$ \\ 
Sphere & 2 & SE-ARD & $[-5.12,5.12]^2$ \\ 
Sum Squares & 2 & SE-ARD & $[-10,10]^2$ \\ 
Trid & 2 & SE-ARD & $[-4,4]^2$ \\ 
Ursem Waves & 2 & SE-ARD & $[-1.2,1.2]\times [-0.9,1.2]$ \\ 
\bottomrule
\caption[Benchmarks]{Benchmark functions in Bayesian optimization experiments.}
\label{tab:benchmarks}
\end{longtable}
 
%\end{longtable}
\clearpage
\end{appendices}

\end{document}